\documentclass[conference]{IEEEtran}
\IEEEoverridecommandlockouts
\usepackage{cite}
\usepackage{amsmath,amssymb,amsfonts}
\usepackage{algorithmic}
\usepackage{graphicx}
\usepackage{textcomp}
\usepackage{xcolor}
\usepackage{multirow}
\def\BibTeX{{\rm B\kern-.05em{\sc i\kern-.025em b}\kern-.08em
    T\kern-.1667em\lower.7ex\hbox{E}\kern-.125emX}}
    \usepackage[left=0.68in,right=0.68in,top=0.75in,bottom=1.1in]{geometry}
\begin{document}

\title{Revising the Problem of Partial Labels from the Perspective of CNNs' Robustness 
}

\author{\IEEEauthorblockN{1\textsuperscript{st} Xin Zhang}
\IEEEauthorblockA{
\textit{University of Southern Maine}\\
Portland, United States \\
xin.zhang@maine.edu}
\and
\IEEEauthorblockN{2\textsuperscript{nd} Yuqi Song}
\IEEEauthorblockA{
\textit{University of Southern Maine}\\
Portland, United States \\
yuqi.song@maine.edu}

\and
\IEEEauthorblockN{3\textsuperscript{rd} Wyatt McCurdy}
\IEEEauthorblockA{
\textit{University of Southern Maine}\\
Portland, United States \\
wyatt.mccurdy@maine.edu}

\and
\IEEEauthorblockN{4\textsuperscript{th} Xiaofeng Wang}
\IEEEauthorblockA{
\textit{University of South Carolina}\\
Columbia, United States \\
wangxi@cec.sc.edu}

\and
\IEEEauthorblockN{5\textsuperscript{th} Fei Zuo}
\IEEEauthorblockA{
\textit{University of Central Oklahoma}\\
Edmond, United States \\
fzuo@uco.edu}

}
\maketitle

\begin{abstract}
Convolutional neural networks (CNNs) have gained increasing popularity and versatility in recent decades, finding applications in diverse domains. 
These remarkable achievements are greatly attributed to the support of extensive datasets with precise labels. However, annotating image datasets is intricate and complex, particularly in the case of multi-label datasets. Hence, the concept of partial-label setting has been proposed to reduce annotation costs, and numerous corresponding solutions have been introduced.
The evaluation methods for these existing solutions have been primarily based on accuracy. That is, their performance is assessed by their predictive accuracy on the test set. However, we insist that such an evaluation is insufficient and one-sided.
On one hand, since the quality of the test set has not been evaluated, the assessment results are unreliable. On the other hand, the partial-label problem may also be raised by undergoing adversarial attacks. Therefore, incorporating robustness into the evaluation system is crucial.
For this purpose, we first propose two attack models to generate multiple partial-label datasets with varying degrees of label missing rates. Subsequently, we introduce a lightweight partial-label solution using pseudo-labeling techniques and a designed loss function. Then, we employ D-Score to analyze both the proposed and existing methods to determine whether they can enhance robustness while improving accuracy. Extensive experimental results demonstrate that while certain methods may improve accuracy, the enhancement in robustness is not significant, and in some cases, it even diminishes. 

\end{abstract}

\begin{IEEEkeywords}
computer vision, multi-label classification, CNN robustness, partial labels
\end{IEEEkeywords}

\section{Introduction}
Convolutional Neural Networks (CNNs), with the ability to extract and learn features automatically from raw data have revolutionized the field of computer vision and have achieved state-of-the-art results on various tasks, including image classification~\cite{lu2007survey}, object detection~\cite{zou2023object}, semantic segmentation~\cite{silberman2012indoor}, and so on~\cite{song2018social,song2021computational,dong2022deepxrd}. Recently, with the rapid development of deep learning techniques, CNNs have also become an indispensable tool for many real-world computer vision applications, such as self-driving cars~\cite{grigorescu2020survey}, security and surveillance systems~\cite{kafedziski2018detection}, and medical diagnosis~\cite{litjens2017survey}.

The tremendous success of CNNs is largely attributed to the support of accurate labeling. However, acquiring precise annotations is quite expensive. To economize on annotation costs, previous endeavors introduced the notion of the 'partial-label setting' and suggested various methodologies to tackle this problem, enabling CNNs to employ only a fraction of the labels during training~\cite{zhang2022effective,shi2018transductive,zhang2023towards}.
After a comprehensive review of the literature on the partial-label problem, we noticed that prior works have primarily assessed their proposed methods based on accuracy alone. We consider it one-sided to conclude the effectiveness of the proposed methods in addressing the partial-label problem solely based on this type of evaluation.
On the one hand, previous works merely demonstrated improved accuracy of their proposed solutions on predefined test sets without evaluating the test sets themselves. Therefore, such evaluation results may not be reliable. On the other hand, besides saving annotation costs, adversarial attacks are one of the reasons for partial-label problems. Extensive prior research has proven that CNNs are vulnerable to adversarial attacks~\cite{qiu2019review}, which is a type of attack used to deteriorate the performance of CNNs targeting datasets, image features, label information, or the models themselves. CNNs with poor robustness often experience significant performance degradation when subjected to adversarial attacks.
Therefore, we insist that analyzing the partial-label problem solely from the perspective of accuracy is one-sided. We also need to analyze it from the standpoint of robustness, that is, analyzing CNN's robustness with respect to label removal.

To conduct an analysis of the partial-label problem from a robustness perspective, we first require datasets that have been subjected to adversarial attacks. Such datasets should consist of training images where only a portion of the labels is known after the adversarial attacks. The current datasets are either fully labeled or partially labeled with a fixed quantity of missing labels, making it challenging to effectively verify how proposed methods are affected by varying degrees of label loss.
For this purpose, we initially propose two attack models: random attacks $\mathcal{R}_p$ and targeted attacks $\mathcal{T}_p$. The former randomly removes $p\%$ of the labels from the images in the training set, irrespective of whether they are positive or negative labels. The latter selectively targets only the positive labels in the training set, removing $p\%$ of the positive labels while preserving all the negative labels.
Furthermore, we introduce a lightweight solution to the partial-label problem. It leverages pseudo-labeling techniques and a well-designed loss function.
Moreover, to evaluate whether our method and existing approaches enhance robustness concerning label removal, besides using the mAP evaluation metric, we also employed the D-Score~\cite{zhang2023d} analysis method to assess the robustness of these methods.

Our Contributions are summarized as follows:
\begin{itemize}
    \item We propose two adversarial attack models targeting image labels: targeted attacks and random attacks. These attack methods selectively remove certain labels, transforming the full-label setting into a partial-label setting. Experimental results demonstrate that this attack effectively reduces the performance of existing STOA methods.
    \item We present a lightweight approach to address the partial-label problem, which is achieved through the utilization of pseudo-labeling techniques and an improved loss function, without the need for additional statistical information or network structures.
    \item The extensive experiments on three large-scale public image datasets (COCO, NUS-WIDE, and Pascal VOC) demonstrate that our method outperforms the STOA methods, both in terms of accuracy (mAP) and robustness (D-Score).
\end{itemize}
The rest of the paper is organized as follows. Section~\ref{sec:related} discusses the related work. Our proposed method is presented in Section~\ref{sec:M}. Section~\ref{sec:exp} shows the experimental settings and results.  Finally, conclusions are drawn in Section~\ref{sec:con}.

\section{Related Works}
\label{sec:related}

\subsection{Partial-label Problems}
The partial-label problem means that for one input image in the training set, only a subset of all the labels for it can be observed and the rest remains unknown during the training process~\cite{zhang2022effective,zhang2023towards}. Addressing this problem is meaningful for saving annotation costs.

A straightforward approach for the partial-label problem is BR~\cite{zhang2018binary}, which decomposes the task into a number of binary classification problems, each for one label. Such an approach encounters many difficulties, mainly due to ignoring correlations between labels. PU-learning is an alternative solution~\cite{li2003learning}, which studies the problem with a small number of positive examples and a large number of unlabeled examples for training. Most methods can be divided into the following three categories: two-step techniques~\cite{liu2002partially}, biased learning~\cite{sellamanickam2011pairwise}, and class prior incorporation~\cite{jiang2008novel}. However, all these methods require that the training data consists of positive and unlabeled examples \cite{bekker2020learning}. Pseudo-label~\cite{zhang2022effective} is another solution. Pseudo-labeling was first proposed in \cite{lee2013pseudo}.  The goal of pseudo-labeling in partial-label problems is to generate pseudo-labels for the unobserved part~\cite{shi2018transductive}. 


\subsection{The Evaluation for CNNs' Robustness}

To evaluate CNNs, researchers have proposed several approaches, which can be divided into two categories. The first category involves introducing the traditional software engineering testing method, mutation testing, to CNNs~\cite{hu2019deepmutation++, humbatova2021deepcrime, shen2018munn}. This approach applies carefully designed mutation operators~\cite{ma2018deepmutation} to the CNN model to generate multiple variants. The higher the number of differences between the predictions of the variant models and the original model, the higher the quality of the test set. However, the score itself remains a black box, and the reasons behind the low quality of the test set are still unknown. 
Additionally, effective methods for selecting and combining mutation operators to detect test set quality remain unexplored~\cite{panichella2021we}.
The second category of approaches is based on neuron coverage~\cite{pei2017deepxplore, feng2020deepgini, yu2019test4deep}. These methods use gradient ascent to solve a joint optimization problem that maximizes both neuron coverage and the number of potentially erroneous behaviors, and eventually generate a set of test inputs~\cite{pei2017deepxplore}. However, as noted in~\cite{harel2020neuron}, higher neuron coverage can lead to fewer defects detected, less natural inputs, and more biased prediction preferences. Therefore, developing effective methods for providing white-box scores for CNNs and proposing methods for enhancing these scores is critical for improving robustness and accuracy of~CNNs.

\section{Methodology}
\label{sec:M}
In this section, we introduce details of our proposed methods, including two adversarial attack models, the solution to the partial-label problem, and the evaluation methods.

\subsection{Simulation of Targeting-label Attack}
\begin{figure*}
    \centering
    \includegraphics[width=0.7\textwidth]{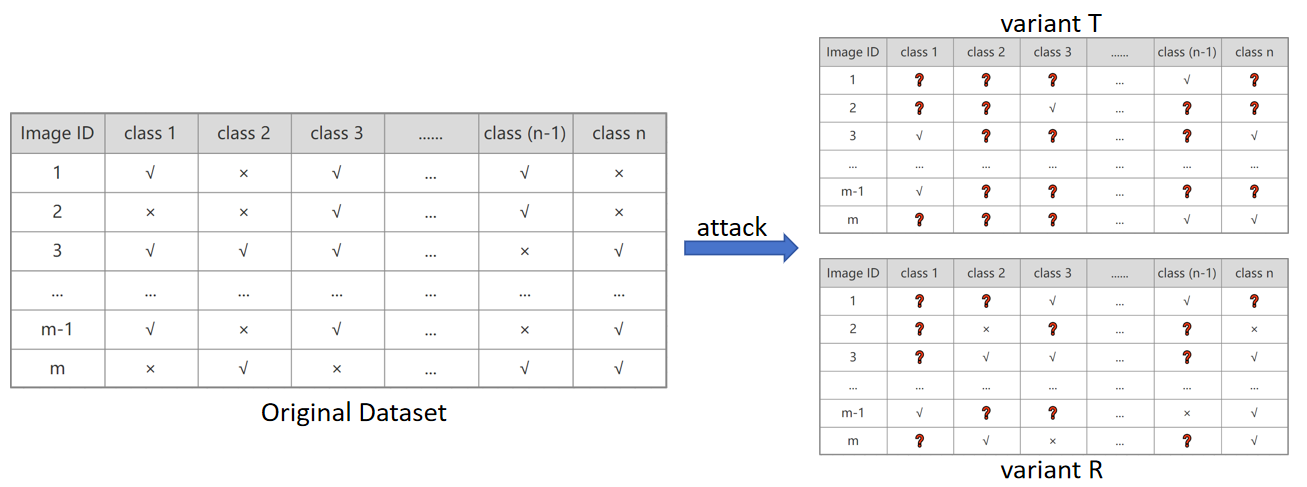}
    \caption{Two attacking models. Variant T is generated under attack $\mathcal{T}$, which removes all negative labels and some positive labels. Variant R, on the other hand, is generated under attack $\mathcal{R}$, where the positive or negative nature of the label is disregarded, and the deletion of labels is entirely random.}
    \label{fig.ams}
\end{figure*}

Label removal is one of the most prevalent adversarial attacks that specifically targets labels. It operates by altering the label distribution through the removal of ground-truth labels, thereby diminishing the model's accuracy and potentially impeding the training process. To validate our method's efficacy in combating adversarial attacks and fortifying the CNN's robustness, we must initially simulate this targeted label attack. We've devised two attack models based on the positive or negative attributes of the targeted labels.
\begin{itemize}
    \item Targeted attacks. In this attack model, we directly eliminate all negative labels and a certain proportion of positive labels. We designate this attack model as $\mathcal{T}_q$, where $q$ represents the deletion percentage of positive labels. Essentially, this attack method removes $\hat{q}$ percent of the labels.:
    $$ \hat{q}\% = \frac{t_n + q\% \times t_p}{t} $$
    where $t$, $t_n$, and $t_p$ stand for the number of total labels, the number of negative labels, and the number of positive labels, respectively. 
    \item Random attacks. In this attack model, we do not distinguish between positive and negative labels; instead, we directly delete labels based on a specific proportion. We denote this attack model as $\mathcal{R}_q$, where $q$ represents the percentage of labels deleted. It is worth noting that this attack model could result in an extreme scenario where all positive labels are removed. This implies that for an image in the training set, its corresponding label contains only negative labels. This situation could easily lead the model to generate a trivial solution, significantly reducing its accuracy. Hence, when designing a solution, it is crucial to address this extreme label imbalance.
\end{itemize}
These two attacking models are summarized in Figure.~\ref{fig.ams}.

\subsection{The Solution for Targeting-label Attack}
To avoid introducing additional computational burden to the model, we propose a lightweight solution that involves modifying only the loss function to enhance the CNN's robustness against targeting-label attacks.

\noindent
\textbf{Pseudo-label.}
We can divide the labels associated with an image into two parts: $E$, which persists after the attack, and $N$, the labels removed due to the attack. For multi-label classification tasks, Binary Cross-Entropy (BCE) commonly serves as the primary loss function, as shown in Equation.~\ref{BCE},
\begin{equation}
    \mathcal{L}_{bce}(\hat{y},y)=-\frac{1}{L}\sum_{i=1}^{L}[(y_i \log(\hat{y}_i)+(1-y_i)\log(1-\hat{y}_i)]\label{BCE}.
\end{equation}
where $y$ and $\hat{y}$ stand for the ground-truth labels and predictions of the classifier, respectively. For part $E$, we compute the loss using the original ground-truth as the target. For the $N$ part, we introduce pseudo-labels and employ them as targets to calculate the loss.
The pseudo-label begins with an initialization of 1, and its value undergoes updates using a historical stack. This stack retains the model's predictions for this label from the previous three epochs. These processes are summarized in Equation.~\ref{equ:pse},
\begin{equation}
    \begin{aligned}
        \widetilde{y}_{i_j} = \begin{cases}
    f(S_{i_j},\alpha,\beta,\gamma), & \textit{update}\\
    1, & \textit{initialization}
    \end{cases}\label{equ:pse}
    \end{aligned}
\end{equation}
where $S_{i_j}$ stands for the historical stack, which always reserves the predictions of the last three epochs for the $i$th image's $j$th category, that is, $size(S_{i_j}) = 3$. $\alpha$, $\beta$, and $\gamma$ stand for the weight for the three elements in $S_{i_j}$ during the calculation of function $f(\cdot)$, 
\begin{equation}
    f(S_{i_j},\alpha,\beta,\gamma) = \alpha S_{i_j}[0] + \beta S_{i_j}[1] + \gamma S_{i_j}[2],
\end{equation}
where $S_{i_j}[k] = \hat{y}_{i_j}^{e-k}$, $e$ represents the index of current epoch number, and $\hat{y}_{i_j}$ stands for the prediction for the $i$th image's $j$th category.
The values of $\alpha$, $\beta$, and $\gamma$ are decided by extensive experiments, and and $\alpha + \beta+\gamma = 1$.

It is essential to note that initializing the pseudo-label as 1 stems from the prevalence of numerous negative labels in image datasets. This often leads to label imbalance, potentially prompting the model to generate trivial solutions, that is, directly predicting each category as negative. The initialization of the pseudo-label as 1 effectively alleviates this issue. While updating the pseudo-label, the historical stack aids in tracking the label value fluctuations over the past three instances, ensuring a smoother update.

\noindent
\textbf{Loss function.}
We employ Binary Cross Entropy (BCE) as our loss function. In the part $E$, we compute the loss value using ground-truth as the target, while for the $N$ part, we calculate the loss value using the pseudo label as the target, as illustrated in Equation.~\ref{baseloss},
\begin{equation}
    \mathcal{L} = \mathcal{L}_{bce}(\hat{y},y) + \mathcal{L}_{bce}(\hat{y},\widetilde{y}) ,\label{baseloss}
\end{equation}
where $y$, $\hat{y}$, and $\widetilde{y}$ stand for the ground-truth labels, the predictions, and the pseudo labels respectively. In Equation.~\ref{baseloss}, the first term represents for the loss value of the $E$ part, and the second term stands for the $N$ part. 
Building upon this, we introduce an attention-shifting parameter $e(\cdot)$ to progressively redirect attention from the part $E$ to the part $N$ during the loss function computation. The rationale behind this design is that at the initial training stages, since pseudo labels are initially set to 1, they may significantly deviate from the actual labels, potentially resulting in unreliable loss value calculations for this part. Therefore, during the early training phase, we aim to focus more on the $E$ part. As training advances and the pseudo labels are continually updated, their reliability gradually increases. Consequently, as training progresses, we gradually shift attention towards the $N$ part.

In previous research, the attention-shifting parameter $e(\cdot)$ has often been applied using a linear function. However, in this context, we employ an exponential function for implementation, as Equation~\ref{efun},
\begin{equation}
    e(n_c, n_t) = e^{n_c - n_t},\label{efun}
\end{equation}
where $n_c$ and $n_t$ stand for the index of the current epoch number and the total epoch numbers respectively. Compared to the linear function, the exponential function exhibits faster changes toward the end of the training process, while its alterations are more gradual during the initial training stages. This approach ensures that the pseudo labels have ample time for updating during training and gradually become dominant in calculating the loss value as training progresses. 

In addition, to prevent the occurrence of trivial solutions, we design an approach for penalizing such outcomes. This approach calculates the difference between the current model's predictions and the trivial solution. When this disparity is tiny, it indicates that the current model has potentially produced a trivial solution. In such cases, we apply a penalty $\mathcal{P}$ to these predictions. We use the L2 norm to compute the difference between the predictions and the trivial solution, as depicted in Equation.~\ref{pen},
\begin{equation}
    \mathcal{P} = 1 - \sqrt{\sum_{i=0}^{m}(\hat{y}_i-\widetilde{y}_i)^2},\label{pen}
\end{equation}
where $m$ stands for the number of existing labels for this image, and because $\sqrt{\sum_{i=0}^{m}(\hat{y}_i-\widetilde{y}_i)^2} \in [0,1]$, $\mathcal{P}$ is always larger than 0.

Combining all these, our final loss function is shown as Equation.~\ref{finalloss},
\begin{equation}
    \mathcal{L} = \mathcal{L}_{bce}(\hat{y},y) + e^{n_c - n_t}\times\mathcal{L}_{bce}(\hat{y},\widetilde{y}) + \mathcal{P},\label{finalloss}
\end{equation}

\subsection{Evaluating the Robustness of Proposed Solutions}
To demonstrate that our proposed solutions can indeed improve the robustness, we adopt D-Score~\cite{zhang2023d} to analyze the presented method and other comparisons. 
D-Score is a quantitative method for analyzing the robustness of CNNs. It analyzes the model's attention distribution and the dataset's feature distribution through the deletion of mutation operators and feature shifting. Subsequently, it evaluates the CNN's robustness by computing the similarity between these two distributions.

\section{Experiments}
\label{sec:exp}
The detailed experimental settings and results are summarized in this section.

\subsection{Datasets}
We conduct comprehensive experiments on three large-scale multi-label image datasets: COCO~\cite{coco}, NUS-WIDE~\cite{nus}, and Pascal VOC~\cite{pascal}.  Each instance in these three datasets is fully annotated with clean labels that can be used as the GT in performance evaluation. 

\subsection{Network Structure and Hyper-Parameters}
Following~\cite{zhang2022effective}, we adopt the same network structure by using an end-to-end network for all experiments: a ResNet-50 \cite{xie2017aggregated}, pre-trained on ImageNet \cite{deng2009imagenet}, as the backbone and a fully connected layer, which is the same as the multi-label classifier under FOL setting. Our approach does not add any extra structure to the network. 

Additionally, we also follow the same training hyperparameter selection to control variables. That is, we train our classifier for 10 epochs, and for the learning rate and batch size, we use a hyperparameter search method and select the hyperparameters with the bast mAP on the validation set, where the learning rate is in $[1e-3, 1e-4, 1e-5, 1e-6]$ and batch size is in $[8,16]$.

\subsection{Experimental Results}

\begin{figure}
    \centering
    \includegraphics[width=0.45\textwidth]{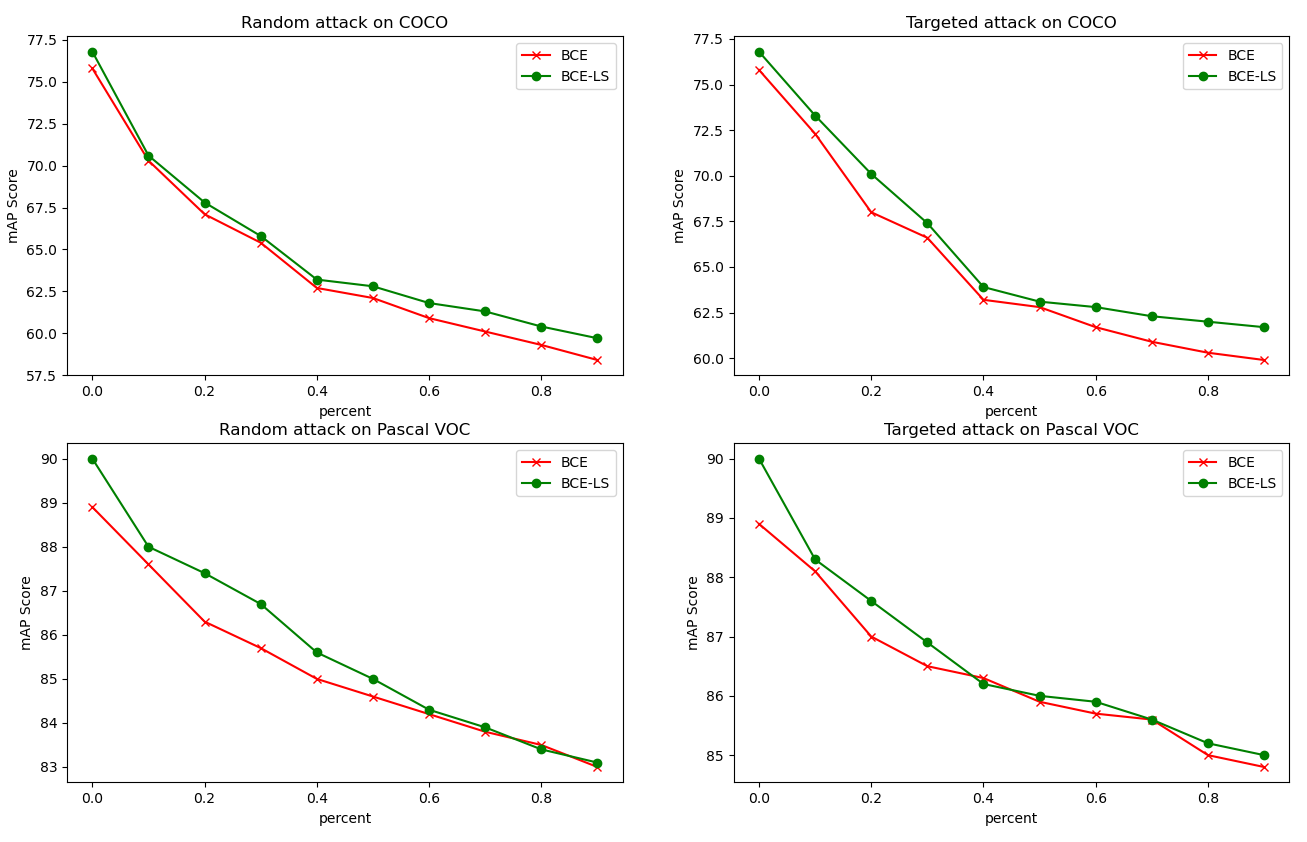}
    \caption{The effectiveness of our attacking models in decreasing the mAP scores. The first row is conducted on the COCO dataset, and the second row is on Pascal VOC.}
    \label{fig:effams}
\end{figure}

\noindent
\textbf{The effectiveness of attacking models.}
To demonstrate the significant threat our proposed attack models pose to CNNs, we opted to assess the performance of two commonly employed solutions for multi-label classification problems: Binary Cross-Entropy (BCE) and BCE with Label Smoothing (BCE-LS). This evaluation was conducted on COCO and Pascal VOC, two extensive image datasets, under varying degrees of attack.
The experimental results are shown in Figure.~\ref{fig:effams}, which fully demonstrate the effectiveness of our proposed attacking models. 

\noindent
\textbf{The effectiveness of our proposed method in addressing the partial-label problem.}
To demonstrate the effectiveness of our approach, we choose several state-of-the-art methods as comparisons, including AN~\cite{kundu2020exploiting}, WAN~\cite{mac2019presence}, ROLE~\cite{cole2021multi}. Then, we conduct targeted attacking and random attacking on three public image datasets, that is, COCO, Pascal VOC and NUS-WIDE, to generate several variants of training sets. 
The experimental results under targeted attacking are summarized in Table.~\ref{table:tm}.
There is also a special variation in this table, namely $\mathcal{T}_s$. This variant is generated through a specific form of targeted attack, wherein for each image, we retain only one positive label and eliminate all other labels. This variant holds particular practical significance: in such scenarios, we only require a single annotation for each image, leading to a substantial reduction in annotation costs. Hence, we specifically examine the performance of our approach concerning this variant.
Table.~\ref{table:rm} summarizes the results by random attacking. These results demonstrate the effectiveness of our proposed method in addressing the partial-label problem and resisting the two proposed attacks.

\begin{table*}[htbp]
\label{table3}
\caption{The mAP results on COCO, Pascal VOC, and NUSWIDE datasets attacked by $\mathcal{T}$. $\mathcal{T}_{0.2}$, $\mathcal{T}_{0.4}$, and $\mathcal{T}_{0.6}$ represent randomly retaining 80\%, 60\% and 40\% of positive labels for each instance, respectively, and discarding all negative labels. In each column, we bold the best-performing method.}
\begin{center}
\begin{tabular}{|c|c|c|c|c|c|c|c|c|c|c|c|c|}
\hline
\multirow{3}*{}&\multicolumn{4}{|c|}{\textbf{COCO}}&\multicolumn{4}{|c|}{\textbf{Pascal VOC}}&\multicolumn{4}{|c|}{\textbf{NUSWIDE}} \\
\hline
 & $\mathcal{T}_s$ & $\mathcal{T}_{0.6}$ &$\mathcal{T}_{0.4}$& $\mathcal{T}_{0.2}$ & $\mathcal{T}_s$ & $\mathcal{T}_{0.6}$ &$\mathcal{T}_{0.4}$& $\mathcal{T}_{0.2}$ & $\mathcal{T}_s$ & $\mathcal{T}_{0.6}$ &$\mathcal{T}_{0.4}$& $\mathcal{T}_{0.2}$\\
\hline
AN \cite{kundu2020exploiting} &63.9 &66.4 &67.0 &69.2 &84.7 &86.8 &87.3 &88.0 &40.0 &45.1 &48.3 &50.3 \\
\hline
WAN \cite{mac2019presence} &64.8 &69.1 &70.8 &71.2 &85.9 &87.5 &87.9 &88.0 &43.7 &46.9 &48.5 &50.6 \\
\hline
ROLE \cite{cole2021multi} &65.9 &73.1 &75.4 &77.1 &87.0 &88.9 &90.0 &90.3 &43.2 &48.0 &50.4 &52.0\\
\hline
Ours & \textbf{67.1}& \textbf{74.0}& \textbf{75.8}& \textbf{77.4}& \textbf{87.2}& \textbf{89.2}& \textbf{90.3}& \textbf{90.8}& \textbf{44.9}& \textbf{48.7} & \textbf{50.9} & \textbf{52.8} \\
\hline
\end{tabular}
\label{table:tm}
\end{center}
\end{table*}

\begin{table*}[htbp]
\label{table3}
\caption{The mAP results on COCO, Pascal VOC, and NUSWIDE datasets attacked by $\mathcal{R}$. $\mathcal{R}_{0.2}$, $\mathcal{R}_{0.4}$, and $\mathcal{R}_{0.6}$ represent randomly retaining 80\%, 60\% and 40\% of labels for each instance, respectively. In each column, we bold the best-performing method.}
\begin{center}
\begin{tabular}{|c|c|c|c|c|c|c|c|c|c|}
\hline
\multirow{3}*{}&\multicolumn{3}{|c|}{\textbf{COCO}}&\multicolumn{3}{|c|}{\textbf{Pascal VOC}}&\multicolumn{3}{|c|}{\textbf{NUSWIDE}} \\
\hline
 & $\mathcal{R}_{0.6}$ &$\mathcal{R}_{0.4}$& $\mathcal{R}_{0.2}$ & $\mathcal{R}_{0.6}$ &$\mathcal{R}_{0.4}$& $\mathcal{R}_{0.2}$ & $\mathcal{R}_{0.6}$ &$\mathcal{R}_{0.4}$& $\mathcal{R}_{0.2}$\\
\hline
AN \cite{kundu2020exploiting} &60.1 & 62.3 & 63.7 & 78.6 & 80.1 & 82.3 & 38.7 & 40.5 & 44.1 \\
\hline
WAN \cite{mac2019presence} & 60.8 & 63.1 & 64.3 & 80.2 & 81.7 & 84.4 & 40.2 & 41.1 & 43.7 \\
\hline
ROLE \cite{cole2021multi} & 61.1 & 63.9 & 65.0 & 81.0 & 83.1 & 84.9 & 40.8 & 41.9 & 44.2 \\
\hline
Ours & \textbf{61.6}&\textbf{64.3} &\textbf{66.0} &\textbf{81.8} &\textbf{85.4} &\textbf{85.9} &\textbf{42.1} &\textbf{42.7} &\textbf{44.7}  \\
\hline
\end{tabular}
\label{table:rm}
\end{center}
\end{table*}

\noindent
\textbf{The analysis of robustness.}
We summarize the results of the D-Score analysis in Table.~\ref{table:dsPPL}.

\begin{table*}[htbp!]
\caption{The mAP, robustness score, fitness score and D-Score results undergoing $\mathcal{T}_{0.4}$ on COCO and Pascal VOC. $\mathcal{T}_{0.4}$ represents randomly retaining 60\% of positive labels for each instance, respectively, and discarding all negative labels. For the robustness score, the smaller the better. As for the mAP, fitness score and D-Score, the higher the better. In each column, we bold the best-performing method and underline the second-best one.}
\begin{center}
\begin{tabular}{|c|c|c|c|c|c|c|c|c|}
\hline
\multirow{2}{*}{}&\multicolumn{4}{c|}{\textbf{COCO $\mathcal{T}_{0.4}$}}&\multicolumn{4}{c|}{\textbf{Pascal VOC $\mathcal{T}_{0.4}$}} \\ \hline
& mAP & robust & fitness & D-Score & mAP & robust & fitness & D-Score \\ \hline
AN~\cite{kundu2020exploiting} & 67.0 & 0.397 & 0.765 & 0.368 & 87.3 & 0.288 & 0.816 & 0.528 \\ \hline
WAN~\cite{mac2019presence}    & 70.8 & 0.416 & 0.777 & 0.361 & 87.9 & 0.306 & 0.831 & 0.525  \\ \hline
Role~\cite{cole2021multi}     & \underline{75.4} & \underline{0.289} & \underline{0.801} & \underline{0.512} & \underline{90.0} & \underline{0.261} & \underline{0.847} & \underline{0.568}   \\ \hline
Ours                          & \textbf{75.8} & \textbf{0.272} & \textbf{0.809} & \textbf{0.537} & \textbf{90.3} & \textbf{0.237} & \textbf{0.852} & \textbf{0.615}   \\ \hline
\end{tabular}
\end{center}
\label{table:dsPPL}
\end{table*}

\section{Conclusion}
\label{sec:con}
The remarkable success of CNNs relies heavily on the support of large, high-quality labeled datasets. However, acquiring such datasets is costly due to the extensive manual annotation involved, particularly in multi-label datasets. To address this challenge, numerous methods have been proposed to train CNNs using partial-label datasets. Yet, the evaluation of these solutions has been limited to accuracy, which we deem insufficient. It is crucial to include robustness in the assessment, as the quality of the test sets used for evaluation remains unverified and the partial-label issue may stem from adversarial attacks, closely linked to CNNs' robustness.
To tackle these challenges, we introduce two adversarial attack models aimed at removing specific labels and generating partial-label datasets. Subsequently, we propose a lightweight solution for partial-label problems using pseudo-label techniques. Finally, we conduct an analysis using D-Score and mAP evaluation metrics to assess both the robustness and accuracy of our proposed method and some state-of-the-art methods. Experimental results demonstrate that while our method significantly enhances accuracy, it also notably improves robustness. Conversely, certain existing methods exhibit improved accuracy but a simultaneous decline in robustness.

\bibliographystyle{IEEEtran}
\bibliography{reference}

\end{document}